\documentclass[11pt]{article}
\usepackage[letterpaper, left=1in, right=1in, top=1in, bottom=1in]{geometry}

\usepackage{amsthm}
\usepackage{mathtools}
\usepackage{amsmath}
\usepackage{bbm}
\usepackage{amsfonts}
\usepackage{amssymb}
\usepackage{multirow}
\usepackage{placeins}

\usepackage[dvipsnames]{xcolor}
\usepackage[colorlinks=true, linkcolor=blue!70!black, citecolor=blue!70!black,urlcolor=black,breaklinks=true]{hyperref}
\usepackage{microtype}
\usepackage{graphicx}
\usepackage{subfigure}
\usepackage{booktabs}
\usepackage{enumitem}
\usepackage{ascii}

\usepackage[capitalize,noabbrev]{cleveref}

\usepackage{natbib}
\theoremstyle{plain}
\newtheorem{theorem}{Theorem}[section]

\newtheorem{corollary}[theorem]{Corollary}
\theoremstyle{definition}

\theoremstyle{remark}

\usepackage[textsize=tiny]{todonotes}

\newcommand{\defeq}{\vcentcolon=}

\DeclareMathOperator*{\Multi}{Multi}

\def\Dir{\text{Dir}}
\def\Var{\text{Var}}
\DeclareMathOperator{\Tr}{Tr}

\def\E{\mathbb E}

\def\a{\mathbf a}

\def\p{\mathbf p}

\def\t{\mathbf t}

\def\x{\mathbf{x}}

\usepackage{parskip}

\input{defs}
\newif\ifshort

\title{Dirichlet  Proportions Model for Hierarchically Coherent Probabilistic Forecasting}

\author{
Abhimanyu Das, Weihao Kong, Biswajit Paria\thanks{Part of this work was done when BP was at Carnegie Mellon University.}, Rajat Sen\\
\small (ordered alphabetically)\\
Google\\
\texttt{\small\{abhidas, weihaokong, biswajitsc, senrajat\}@google.com}
}
\date{}

\begin{document}
\maketitle

\begin{abstract}
Probabilistic, hierarchically coherent forecasting is a key problem in many practical forecasting applications -- the goal is to obtain coherent probabilistic predictions for a large number of  time series  arranged in a pre-specified tree hierarchy.  In this paper, we present an end-to-end deep  probabilistic model for hierarchical forecasting that is motivated by a classical top-down strategy. It jointly learns the distribution of the root time series, and the (dirichlet) proportions according to which each parent time-series is split among its children at any point in time. 
The resulting forecasts are naturally coherent, and provide probabilistic predictions over all time series in the hierarchy. We experiment on several public datasets and demonstrate significant improvements of up to 26\% on most datasets compared to state-of-the-art baselines. Finally, we also provide theoretical justification for the superiority of our top-down approach compared to the more traditional bottom-up modeling.
\end{abstract}

\section{Introduction}
\label{sec:intro}
A central problem in multivariate forecasting is the need to forecast a large group of time series arranged in a natural hierarchical structure, such that time series at higher levels of the hierarchy are aggregates of time series at lower levels.
For example, hierarchical time series are common in retail forecasting applications~\citep{fildes2019retail}, where the time series may capture retail sales of a company at different granularities such as item-level sales, category-level sales, and department-level sales. 
In electricity demand forecasting~\citep{van2015game}, the time series may correspond to electricity consumption at different granularities, starting with individual households, which could be progressively grouped into city-level, and then state-level consumption time-series.  The hierarchical structure among the time series is usually represented as a tree, with leaf-level nodes corresponding to time series at the finest granularity, while higher-level nodes represent coarser-granularities and are obtained by aggregating the values from its children nodes.

Since  businesses usually require forecasts at various different granularities, 
the goal is to obtain accurate forecasts for time series at every level of the hierarchy.
Furthermore, to ensure decision-making at different hierarchical levels are aligned, it is essential to generate predictions that are  \emph{coherent} \citep{hyndman2011optimal} with respect to the hierarchy, that is, the forecasts of a parent time-series should be equal to the sum of forecasts of its children time-series. For example, the sum of the sales predictions for each shoe style should equal the sales prediction for the shoe category~\footnote{Note that this is a non-trivial constraint. For example, generating independent predictions for each time series in the hierarchy using a standard multivariate forecasting model does not guarantee coherent predictions.}.
Finally, to facilitate better decision making, there is an increasing shift towards probabilistic forecasting~\citep{berrocal2010probabilistic, gneiting2014probabilistic}; that is, the forecasting model should quantify the uncertainty in the output and produce a probability distribution, instead of a single point estimate, for predictions.

In this paper, we address the problem of obtaining coherent probabilistic forecasts for large-scale hierarchical time series. 
While there has been a plethora of work on multivariate forecasting, there is significantly limited research on multivariate forecasting for hierarchical time series that satisfy the requirements of both  hierarchical coherence {\bf and} probabilistic predictions.

There are numerous recent works on deep neural network-based multivariate forecasting \citep{salinas2020deepar, oreshkin2019n, rangapuram2018deep, benidis2020neural, sen2019think, olivares2022nbeatsx}, including probabilistic multivariate forecasting~\citep{salinas2019high,rasul2021multivariate} and even graph neural network(GNN)-based models for forecasting on time series with graph-structure correlations~\citep{bai2020adaptive,cao2021spectral,yu2017spatio,li2017diffusion}. However, none of these works ensure coherent predictions for hierarchical time series.

On the other hand, several papers specifically address hierarchically-coherent forecasting ~\citep{hyndman2016fast, taieb2017coherent, van2015game, hyndman2016fast, ben2019regularized, wickramasuriya2015forecasting, wickramasuriya2020optimal, mancuso2021machine, abolghasemi2019machine}, based on the idea of \textit{reconciliation}.
This involves a two-stage process where the first stage generates independent (possibly incoherent) univariate base forecasts, and is followed by a second \textit{reconciliation} stage that adjusts these forecasts using the hierarchy structure, to finally obtain coherent predictions. These approaches are usually disadvantaged in terms of using the hierarchical constraints only as a post-processing step, and not during generation of the base forecasts.
Furthermore,
with the exception of \cite{taieb2017coherent}, which is a two-stage reconciliation-based  model for coherent probabilistic hierarchical forecasting, 
most of these approaches cannot directly handle probabilistic forecasts. 

More recently, there has been some work(~\cite{rangapuram2021end,han2021simultaneously}) that propose single-stage, end-to-end deep neural architectures that directly produce hierarchically-coherent (or approximately coherent) probabilistic forecasts without a need for a post-processing step. 

 
In this paper, we present an alternate approach to  end-to-end deep probabilistic forecasting for hierarchical time series, motivated by a classical method that has not received much recent attention: top-down forecasting. The basic idea is to first model the top-level forecast in the hierarchy tree, along with the ratios or proportions according to how the top level forecasts should be distributed among the children time-series in the hierarchy. The resulting predictions are naturally coherent.  Early top-down approaches were non-probabilistic, and were rather simplistic in terms of modeling the proportions; for example,  by separately modeling the top-level forecast, and then deriving the proportions from historical averages~\citep{gross1990disaggregation}, or from independently generated (incoherent) forecasts of each time-series from another model~\citep{athanasopoulos2009hierarchical}. In this paper, we showcase how modeling both the top-level forecast and the proportions jointly through a single-stage deep probabilistic model can obtain state-of-the-art results for probabilistic, hierarchically-coherent forecasts. 

Crucially, our  proposed model (and indeed all top-down approaches for forecasting) relies on the intuition that the top level time series in a hierarchy is usually much less noisy and less sparse, and hence much easier to predict. Furthermore, it might be easier to predict proportions (that are akin to scale-free normalized time-series) at the lower level nodes than the actual time series themselves. 

Our approach involves learning a single end-to-end deep model to jointly model ``families``, consisting of a parent and its children timeseries, and predict both the parent timeseries and the proportions along which it is disaggregated among its children. We use a Dirichlet distribution~\citep{olkin1964multivariate} to model the distribution of proportions for each parent-children family in the hierarchy. The parameters of the Dirichlet distribution for each family is obtained from an MLP (Multi Layer Perceptron) based encoder-decoder model  with multi-headed self-attention~\citep{vaswani2017attention}, that is jointly learnt from the whole dataset.


We validate our model against state-of-the art probabilistic hierarchical forecasting baselines on six public datasets, and demonstrate  significant gains using our approach, outperforming the  baselines on most datasets with improvements of up to 26\% in terms of CRPS scores. 

Additionally, we theoretically analyze the advantage of the top-down approach (over a bottom-up approach) in a simplified regression setting for hierarchical prediction, and thereby provide theoretical justification for our top-down model. Specifically, we prove that for a $2$-level hierarchy of $d$-dimensional linear regression with a single root node and $K$ children nodes, the excess risk of the bottom-up approach is $\min(K, d)$ time bigger than the one of the top-down approach in the worst case. This validates our intuition that it is easier to predict proportions than the actual values.
\section{Background}
\label{sec:psetting}
Hierarchical forecasting is a multivariate forecasting problem, where we are given a set of $N$ univariate time-series (each having $T$ time points) that satisfy linear aggregation constraints specified by a predefined hierarchy. More specifically,  
the data can be represented by a matrix $\bY \in \reals^{T \times N}$, where $\by^{(i)}$ denotes the $T$ values of the $i$-th time series, $\by_t$ denotes the values of all $N$ time series at time $t$, and $y_{t,i}$ the value of the $i$-th time series at time $t$. We will assume that $y_{i, t} \geq 0$, which is usually the case in all retail demand forecasting datasets (and is indeed the case in all public hierarchical forecasting benchmarks~\citep{wickramasuriya2015forecasting}). 
We compactly denote the $H$-step history of $\bY$ by $\bY_\cH = [\by_{t-H}, \cdots, \by_{t-1}]^\top \in \reals^{H \times N}$ and the $H$-step history of $\by^{(i)}$ by $\by_\cH^{(i)} = [\by_{t-H}\idx{i}, \cdots, \by_{t-1}\idx{i}] \in \reals^H$. Similarly we can define the $F$-step future as $\bY_\cF = [\by_t, \cdots, \by_{t+F-1}]^\top \in \reals^{F \times N}$. We use the $\hat{\cdot}$ notation to denote predicted values, and the ${\cdot}^\top$ notation to denote the transpose. We denote the matrix of external covariates like holidays etc by $\bX \in \reals^{T\times D}$, where the $t$-th row denotes the $D$-dimensional feature vector at the $t$-th time step. For simplicity, we assume that the features are shared across all time series\footnote{Note that our modeling can handle both shared and time-series specific covariates in practice.}. We define $\bX_{\cH}$ and $\bX_\cF$ similarly. The $\hat{\cdot}$ notation will be used to denote predicted values, for instance $\hat{\bY}_{\cF}$ denotes the prediction in the future. We will also use numpy tensor notation i.e $\bX[i:j, r:c]$ would denote the sub-matrix in rows $\{i, i+1, \cdots, j-1\}$ and columns $\{r, r+ 1, \cdots c-1\}$. Further, using $:$ would denote selecting all rows (or columns) depending on the axis.

\begin{figure}[ht!]
    \centering
    \includegraphics[width=\columnwidth]{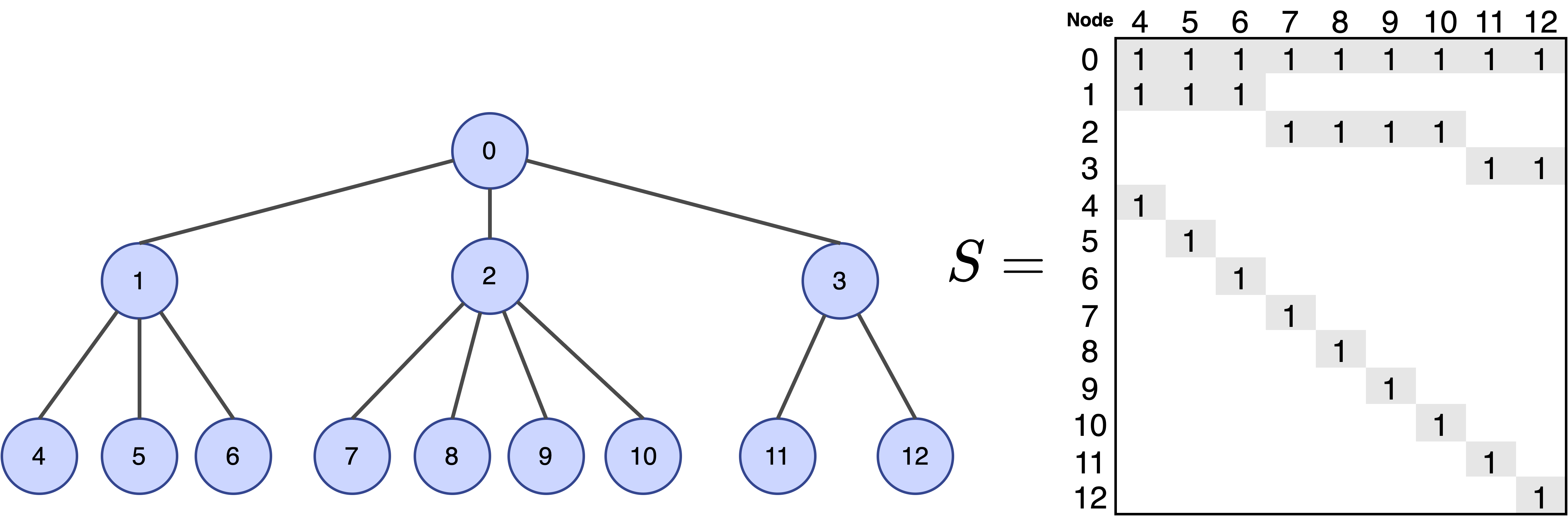}\hfill
    \caption{\small An example of a hierarchy and its corresponding $\bS$ matrix. The rows and columns of the matrix are indexed by the corresponding nodes for easier interpretation. The empty cells of the matrix are zeros, and hence omitted from the figure.}
    \label{fig:hierarchy}
\end{figure}

{\bf Hierarchy.~}
The $N$ time series are arranged in a tree hierarchy, with $m$ leaf time-series, and $k=N-m$ non-leaf (or aggregated) time-series that can be expressed as the sum of its children time-series, or alternatively, the sum of the leaf time series in its sub-tree. Let $\boldb_t \in \reals^{m}$ be the values of the $m$ leaf time series at time $t$, and $\br_t \in \reals^{k}$ be the values of the $k$ aggregated time series at time $t$. The hierarchy is encoded as an aggregation matrix $\bR \in \{0,1\}^{k \times m}$ , where an entry $R_{ij}$ is equal to 1 if the $i$-th aggregated time series is an ancestor of the $j$-th leaf time series in the hierarchy tree, and 0 otherwise. 
We therefore have the aggregation constraints $\br_t = \bR\boldb_t$ or $\by_t=[\br_t^\top ~~ \boldb_t^\top]^\top = \bS\boldb_t$ where $\bS^T = [\bR^\top \vert \bI_m]$. Here, $\bI_m$ is the $m \times m$ identity matrix. 
Such a hierarchical structure is ubiquitous in multivariate time series from many domains such as retail, traffic, etc, as discussed earlier. We provide an example tree with its $\bS$ matrix in Figure~\ref{fig:hierarchy}. We can extend this equation to the matrix $\bY \in \reals^{T \times N}$. Let $\bB := [\boldb_1; \cdots ;\boldb_m]^\top$ be the corresponding leaf time-series values arranged in a $T \times m$ matrix. Then coherence property of $\bY$ implies that 
$\bY^\top = \bS \bB^\top$. Note that we will use $\bB_{\cF}$ to denote the leaf-time series matrix corresponding to the future time-series in $\bY_{\cF}$.

\textbf{Coherency.~}
Clearly, an important property of hierarchical forecasting is that the forecasts also satisfy the hierarchical constraints 
$\hat{\bY}^\top_{\cF} = \bS \hat{\bB}^\top_{\cF}$. This is known as the \emph{coherence} property which has been used in several prior works~\citep{hyndman2018forecasting, taieb2017coherent}. Imposing the coherence property makes sense since the ground truth data $\bY$ is coherent by construction. Coherence is also critical for consistent decision making at different granularities of the hierarchy. 

Our \emph{objective} is to accurately predict the distribution of the future values $\hat{\bY}_\cF \sim \hat{f}(\bY_\cF)$ conditioned on the history such that any sample $\hat{\bY} \in \reals^{F \times N}$ from the predicted distribution $\hat{f}(\bY_\cF)$ satisfies the coherence property. In particular, $\hat{f}$ denotes the density function (multi-variate) of the future values $\bY_\cF$ conditioned on the historical data $\bX_\cH, \bY_\cH$ and the future features $\bX_\cF$. We omit the conditioning from the expression for readability.

The related work can be broadly classified into (i) Coherent point forecasting that includes but is not limited to approaches like~\citep{hyndman2011optimal, van2015game, panagiotelis2020probabilistic}; (ii) Coherent probabilistic forecasting that includes among others~\citep{taieb2017coherent, rangapuram2021end, olivares2021probabilistic} and (iii) Approximately coherent methods like~\citep{mishchenko2019self, gleason2020forecasting, han2021simultaneously, han2021mecats, paria2021hierarchically}. We include a detailed discussion of related literature in Appendix~\ref{app:rwork}.

\section{The Model}
\label{sec:model}
The basic input data-structure to our model is a \textit{family} $(p, \cL(p))$, where $p$ is a parent node in the hierarchical tree (any non-leaf node) and $\cL(p)$ are its children nodes. In Figure~\ref{fig:hierarchy}, the set of nodes $(2, [7, 8, 9, 10])$ denotes a family. 

Our main contribution is a shared proportions model that takes in the past data points of a family and predicts (i) the future proportions of the children i.e the fractions by which the parent time-series diaggregates into the children time-series in the future (ii) the future values of the parent. The model is designed to capture dependencies among the children through appropriate applications of attention layers and also propagate information between the parent and the children. We train a single shared global proportions model for all the families in the tree.

{\bf Modeling Proportions.} Before we describe the model for forecasting the proportions, we need to formally define the children proportions. For a family $(p, \cL(p))$ let us define the proportions,
\begin{align}
    a_{s,c} = \frac{y_{s, c}}{\sum_{j \in \cL(p)} y_{s, j}}, \quad \text{for all } s \in [T], c\in \cL(p) \label{eq:prop}.
\end{align}
The  matrix $\bA(p) \in \reals^{T \times C}$ denotes the proportions of the children over time, where $C := |\cL(p)|$. We will drop the $p$ in braces when it is clear from context that we are dealing with a particular family $(p, \cL(p))$. As in Section~\ref{sec:psetting}, we use $\bA_\cH$ and $\bA_\cF$ to denote the historical and future proportions. Also, $\ba^{(i)}_{\cH}$ will denote the historical proportions for child $i \in \cL(p)$ and a similar definition holds for $\ba^{(i)}_{\cF}$.

We are interested in predicting the distribution of $\bA_\cF$ given  historical proportions $\bA_\cH$, the parent's history $\by\idx{p}_\cH$ and covariates $\bX$. Note that for any $s\in [T]$, the $s$th row of $\bA(p)$, $\ba_s \in \Delta^{C-1}$ (denotes the $(C-1)$-dimensional simplex). Therefore, our predicted distribution should also be a distribution over the simplex for each row. Hence, we use the Dirichlet~\citep{olkin1964multivariate} family to model the output distribution for each row of the predicted proportions, as detailed later.

We choose the simplest possible architectural building blocks for the task at hand: (i) we use a simple MLP (Multi-Layer Perceptron)  encoder-decoder model for capturing the temporal dependencies in the proportions of a child (independently of other children) (ii) we use multi-headed attention~\citep{vaswani2017attention} to capture dependencies among the children of the family. 

{\it Encoder:~} We first form an encoding depending on the past for each child,
\begin{align*}
 \be_i \leftarrow \enc \left(\ba_{\cH}^{(i)}, \by\idx{p}_\cH, \bX\right)   
\end{align*}
where $\be_i \in \reals^{d_{E}}$ and $d_{E}$ is the encoding dimension. We represent all the children embeddings in the matrix $\bE \in \reals^{d_{E} \times C}$ such that the $i$-th column of $\bE$ is the encoding $\be_i$ of the $i$-th child. The MLP encoder, $\enc(\cdot)$ is applied independently for each child. Each child's embedding can also depend on the past of the parent $\by\idx{p}_\cH$ and the covariates $\bX$; thereby drawing information from higher level time-series.

{\it Attention:~} Then we apply multi-headed attention to mix the encoded information across the children. The input to the attention is $\bE' = [\bE; \mathbf{1}] \in \reals^{d_E \times (C + 1)}$ i.e a dummy column added to the children embeddings (the value of that column will become clear later when we dicuss the parent prediction module). The attention layer is denoted by,
\begin{align}
   \bM'  \leftarrow \att_{g,l} \left(\bE'\right), \label{eq:att_layer}
\end{align}
where $g$ denotes the number of attention heads and $l$ denotes the number of attention layers. Each attention layer is followed by a fully connected layer with ReLU activation and also equipped with a residual connection (similar to the original model in~\citep{vaswani2017attention}). Note that the attention is only applied across the second dimension i.e across the children. The resulting $\bM'$ is in $\reals^{d_A \times (C+1)}$ where $d_A$ is the value dimension in the attention layers. Let $\bM = \bM'[:, 0:C+1]$.

{\it Decoder:~} Now that we have mixed the encoded information among the children, we are ready to decode to obtain the predicted distribution of future proportions of the children. The decoding follows the equations:
\begin{align*}
    \bD &= \deci (\bM), ~~\text{(Output shape: $d_D*F \times C$)},\\
    \bD_{\cF} &= \text{(Reshape of $\bD$ into $d_D \times F \times C$)}, \\
    \hat{a}_{s, c} &= \decf (\bD_{\cF}[:, s, c], \bX_{\cF}[s, :]), ~~\text{(Output shape: 1)} \\
    &\hat{\bA}_{\cF}~\text{is a matrix s.t}~ \hat{\bA}_{\cF}[s, c] = \exp(a_{s, c}).
\end{align*}

In the first equation $\deci(\cdot)$ is a MLP decoder with output dimensions $d_D*F$ that is applied on the first axis of $\bM$. Then the output is reshaped into $\bD_{\cF}$ such that we have a decoded feature of length $d_D$ for all $F$ future time-points and all $C$ children in the family. Then we apply another MLP layer with $\decf(\cdot)$ with output dimension $1$. For each future time-point $s \in [F]$ and children $c \in C$, $\decf$ is applied independently on the concatenation of $\bD_{\cF}[:, s, c]$ and $\bX_{\cF}[s, :]$ to produce the output parameter $\hat{a}_{s, c}$. Intuitively, this final MLP combines the information in the $d_D$ dimensional decoded feature for the children $c$ at future time $s$ along with the future covariates at that time-point to produce the final output. The output for all future time-points and all children are then collected in the matrix $\hat{\bA}_{\cF} \in \reals^{F \times C}$ after passing through the exp. function to ensure positivity.

{\it Loss Function:} We would like to output the predicted distribution of the proportions of the children in the family for all future time-steps. Our loss function is designed such that the output from the preceding decoder step $\hat{\bA}_{\cF}$ can represent such a distribution. Recall that the predicted proportions distributions $\hat{f}(\bA_\cF)$ have to be over the simplex $\Delta^{C-1}$ for each time-point. Therefore we model it by the Dirichlet distribution. In fact our final model output $\hat{\bA}_{\cF}$ represents the parameters of predictive Dirichlet distributions. Specifically, we minimize the loss
\begin{align}
    \ell_c(\bA_\cF, \hat{\bA}_{\cF}) = -\frac{1}{F}\sum_{s=1}^{F} \dmle \left(\ba_s + \epsilon; \hat{\mathbf{a}}_s \right), \label{eq:children_loss}
\end{align}
where $\ba_s$ is the proportion of the children nodes in the family at time $s$ as defined in Equation~\eqref{eq:prop}.
$\dmle(\ba; \vbeta)$ denotes the log-likelihood of Dirichlet distribution for target $\ba$ and parameters $\vbeta$.
\begin{align}
\label{eq:dmle}
    \dmle(\ba; \vbeta) := \sum_{i} (\beta_i - 1) \log (a_i) - \log B(\vbeta),
\end{align}
where $B(\vbeta)$ is the normalization constant. In Eq.~\eqref{eq:children_loss}, we add a small $\epsilon$ to avoid undefined values when the target proportion for some children are zero. Here, $B(\vbeta) = \prod_{i=1}^{C} \Gamma(\beta_i) / \Gamma(\sum_i \beta_i)$ where $\Gamma(.)$ is the well-known Gamma function that is differentiable. In practice, we use Tensorflow Probability~\citep{dillon2017tensorflow} to optimize the above loss function.

{\bf Modeling the Parent.} The remaining task is to predict the future values of the parent node in a family. Recall that the output of the attention layer $\bM'$ in Eq.~\eqref{eq:att_layer} has an extra dimension in the second axis. We will use the output in that dimension as an input to the decoder that predicts the future of the parent. This allows us to distill historical information from the children that might also be useful for predicting the future values of the parent. The decoding for the parent prediction comprises of the following equations:
\begin{align*}
    \bP &= \decip(\bM[:, -1], \by\idx{p}_\cH)~~\text{(Output shape: $d_D*F$)}, \\
    \bP_{\cF} &= \text{(Reshape of $\bP$ into $F \times d_D$)}, \\
    \hat{\bp}_{s} &= \decfp (\bP_{\cF}[s, :], \bX_{\cF}[s, :]), ~~\text{(Output shape: 2)} \\
    &\hat{\bP}_{\cF}~\text{is a matrix s.t}~ \hat{\bp}[s, :] = \hat{\bp}_{s}.
\end{align*}
In the first equation we use the MLP decoder to map the last dimension of the attention encoding from the children along with the past of the parent, to the future decoding of shape $d_D*F$. The decoding is then reshaped to have shape $F \times d_D$. The final parent decoding layer is an MLP $\decfp$ with output dimension $2$ that maps each future time-step's decoded features $\bP_{\cF}[s, :]$ along with the covariates at that time-step to the final predicted parameters for that time-step.

\begin{figure*}
    \centering
    \includegraphics[width=0.77\textwidth]{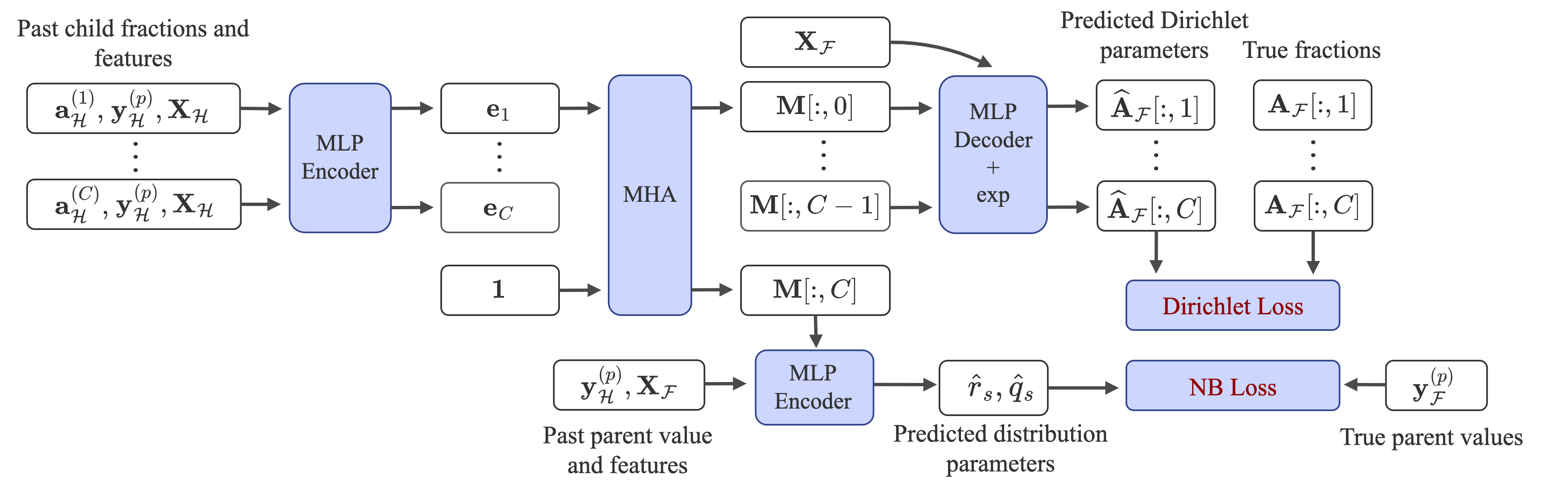}%
    \includegraphics[width=0.23\textwidth]{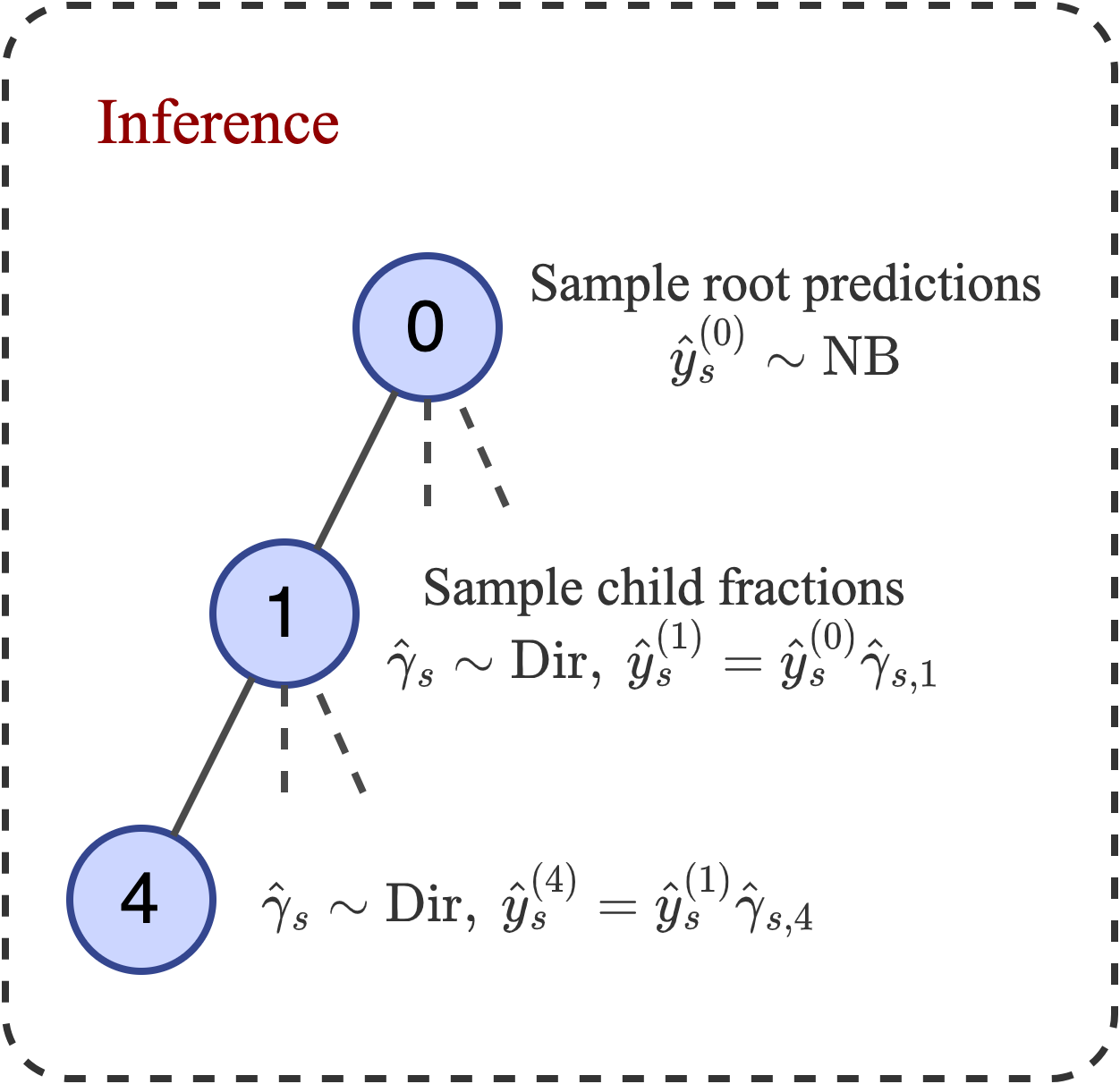}
    \caption{We provide a complete description of the training architecture. MHA denotes multi-head self attention layers. The indices $\{1, \cdots, C\}$ are used to denote the indices of the children of the parent node $p$. During inference, the top level predictions are first sampled from the negative binomial distribution with the predicted parameters $\hat{q}_s, \hat{r}_s$. The children prediction are then successively sampled by first sampling the fractions $\hat{\gamma}_s$ from the Dirichlet distribution and then multiplying with the parent sample.}
    \label{fig:arch}
\end{figure*}

{\it Loss Function.} Since we are interested in probabilistic forecasting, we would like to predict the distribution of the parent's future values. Therefore we will map the output parameters in each future time step ($\hat{\bp}_{s}$ for time-step $s \in [F]$) to the parameters of a negative-binomial distribution. The p.m.f of negative binomial distribution with total count $r > 0$ and success probability $q \in [0, 1]$ is given by,
\begin{equation}
\nble(k; r, q) = \frac{\Gamma(k + r)}{\Gamma(k + 1)\Gamma(r)} (1- q)^k q^r~\text{for}~ k=1,2, \cdots
\end{equation}
We first map the predicted parameters from the parent decoder $\hat{\bp}_{s}$ to the distribution parameters $\hat{r}_s, \hat{q}_s$ using link functions (designed to keep $(r, q)$ in valid ranges):
\begin{align*}
    \hat{r}_s &= \sigma(\hat{\bp}_{s}[0]) \\
    \hat{q}_s &= \frac{1}{1 + \sigma(\hat{\bp}_{s}[1])},
\end{align*}
where $\sigma(x) := (2 + x - |x|)/(2 -x + |x|)$. Thus $\sigma$ maps $\reals \rightarrow \reals^+$. Finally we minimize the negative log likelihood of observing the actual future values of the parents under the negative binomial distribution formed by the predicted parameters,
\begin{align}
    \ell_p(\by^{(p)}_{\cF}, \hat{\bP}_\cF) = - \frac{1}{F} \sum_{s=1}^{F} \log \left( \nble (y_{p, s}; \hat{r}_s, \hat{q}_s ) \right). \label{eq:parent_loss}
\end{align}
We use the negative binomial distribution since a most hierarchical forecasting datasets contain count data that is positive. In such cases, the negative binomial loss has been used with great success in the context of time-series forecasting~\citep{awasthibenefits, salinas2020deepar}.

The final loss is the summation of the children loss in Eq.~\eqref{eq:children_loss} and the parent loss in Eq.~\eqref{eq:parent_loss}. We provide a full illustration of our model in Figure~\ref{fig:arch}. We refer to our model as \oursp.

{\bf Training.} We train our  model  with mini-batch gradient descent where each batch corresponds to different history and future time-intervals of the \textit{same family}. For example, if the time batch-size is $b$ and we are given a family $(p, \cL(p))$ the input proportions that are fed into the model are of shape $b \times H \times C$ and the output distribution parameters for the Dirchlet loss are of shape $b \times F \times C$, where $C = |\cL(p)|$. Note that we only need to load all the time-series of a given family into a batch.

{\bf Inference.} At inference we have to output a representation of the predicted cumulative distribution $\hat{F}(\bY_\cF)$ such that the samples are reconciled as in Section~\ref{sec:psetting}. For ease of illustration, we will demonstrate the procedure for one time point $s \in \{1, \cdots, F\}$. 

We first sample $\hat{y}_s\idx{p}$ from the predictive distribution of the parent for the family containing the root node. For instance, in the tree of Figure~\ref{fig:hierarchy}, this family would be $(0, [1, 2, 3])$. Then for each family $(p, \cL(p))$ in the tree we generate a sample from the Dirchlet distribution with parameters $\hat{\bA}_{\cF}[s, :]$ that represents a sample of the predicted children proportions for that family. The proportion samples and the root sample can be combined to form a reconciled forecast sample $\hat{\by}_s$. We can generate many such samples and then take empirical statistics to form the predictive distribution $\hat{f}(\bY_s)$, which is by definition reconciled.

\section{Theoretical justification for the top-down approach}
\label{sec:theory}
In this section, we theoretically analyze the advantage of the top-down approach over the bottom-up approach for hierarchical prediction in a simplified setting. Again, the intuition is that the root level time series is much less noisy and hence much easier to predict, and it is easier to predict proportions at the children nodes than the actual values themselves. As a result, combining the root level prediction with the proportions prediction actually yields a much better prediction for the children nodes. Consider a 2-level hierarchy of linear regression problem consisting of a single root node (indexed by $0$) with $K$ children. For each time step $t\in [n]$, a global covariate $\x_t\in \reals^d$ is independently drawn from a Gaussian distribution $\x_t \sim \mathcal{N}(0, \Sigma)$, and the value for each node is defined as follows:
\begin{itemize}[noitemsep,topsep=0pt,parsep=0pt,partopsep=0pt,leftmargin=*]
    \item The value of the root node at time $t$ is $y_{t,0}=\theta_0^\top\x_t+\eta_t$, where $\eta_t\in \reals$ is independent of $\x_t$, and satisfies $\E[\eta_t]=0, \text{Var}[\eta_t] = \sigma^2$.
    \item A random $K$-dimensional vector $\a_t\in \reals^K$ is independently drawn from distribution $P$ such that $\E[a_{t,i}] = p_i$ and $\text{Var}[a_{t,i}] = s_i$, where $a_{t,i}$ is the $i$-th coordinate of $\a_t$. For the $i$-th child node, the value of the node is defined as $a_{t,i}\cdot y_{t,0}.$
\end{itemize}
Notice that for the $i$-th child node, $\E[y_{t, i}|\x_t] = p_{i}\theta_0^\top\x_t$, and therefore the $i$-th child node follows from a linear model with coefficients $\theta_i \defeq p_i\theta_0$.

Now we describe the bottom-up approach and top-down approach and analyze the expected excess risk of them respectively. In the bottom-up approach, we learn a separate linear predictor for each child node seprately. 
For the $i$-th child node, the ordinary least square (OLS) estimator is 
$$
\hat\theta^{\mathbf b}_i = \bb{\sum_{t=1}^n \x_t\x_t^\top}^{-1}\sum_{t=1}^n \x_t y_{t,i},
$$
and the prediction of the root node is simply the summation of all the children nodes. 

In the top-down approach, a single OLS linear predictor 
is first learnt for the root node:
$$
\hat\theta_0^{\t} = \bb{\sum_{t=1}^n\x_t\x_t^\top}^{-1}\sum_{t=1}^n\x_ty_{t,0}
$$
Then the proportion coefficient $\hat{p}_i, i\in [K]$ is learnt for each node separately as $\hat{p}_i = \frac{1}{n}\sum_{t=1}^ny_{t,i}/y_{t,0}$ and the final linear predictor for the $i$th child is $\hat{\theta}^{\t}_i = \hat{p}_i\hat\theta_0^{\t}$. Let us define the excess risk of an estimator $\hat\theta_i$ as $r(\hat\theta_i) = (\hat\theta_i - \theta_i)^\top\Sigma (\hat\theta_i - \theta_i)$. The expected excess risk of both approaches are summarized in the following theorem, proved in Appendix~\ref{sec:proof-risk-ratio}.
\begin{theorem}[Expected excess risk comparison between top-down and bottom-up approaches]\label{thm:risk-ratio}
The total expected excess risk of the bottom-up approach for all the children nodes satisfies
$$
\sum_{i=1}^K \E[r(\hat\theta^{\mathbf b}_i) ]\ge \sum_{i=1}^K(s_i +p_i^2) \frac{d}{n-d-1} \sigma^2,
$$ 
and the total expected excess risk of the top-down approach satisfies
\begin{align*}
&\sum_{i=1}^K \E[r(\hat\theta^{\mathbf t}_i) ]\\
&=\frac{\sum_{i=1}^K{s_i}}{n}\theta_0^\top\Sigma\theta_0+\bb{\frac{\sum_{i=1}^K s_i}{n}+\sum_{i=1}^Kp_i^2}\frac{d}{n-d-1}\sigma^2,
\end{align*}
\end{theorem}
Applying the theorem to the case where the proportion distribution $\a_t$ is drawn from a uniform Dirichlet distribution, we show the excess risk of the traditional bottom-up approach is $\min(K,d)$ times bigger than our proposed top-down approach in the following corollary. A proof of the corollary can be found in Appendix~\ref{sec:proof-diri-risk-ratio}
\begin{corollary}\label{cor:diri-risk-ratio}
Assuming that for each time-step $t\in [n] $, the proportion coefficient $\a_t$ is drawn from a $K$-dimensional Dirichlet distribution $\Dir(\alpha)$ with $\alpha_i=\frac{1}{K}$ for all $i\in [K]$ and $\theta_0^\top\Sigma\theta_0=\sigma^2$, then
$$
\frac{\E[\sum_{i=1}^K r(\hat\theta^{\mathbf b}_i)]}{\E[\sum_{i=1}^K r(\hat\theta^{\t}_i)]} = \Omega(\min(K, d)).
$$
\end{corollary}

In Section~\ref{sec:expt}, we show that even the basic topdown approach analyzed in this section outperforms several state of the art methods, thus conforming to our theoretical justification. Our learnt top-down model is a further improvement over the historical fractions.

\section{Experiments}
\label{sec:expt}
We implement our model in Tensorflow~\citep{abadi2016tensorflow} and compare our approach with state of the art models for coherent probabilistic forecasting on 6 hierarchical forecasting datasets. We now describe the datasets along with the corresponding forecasting setups.

\textbf{Datasets.} We experiment with two retail forecasting datasets, M5~\citep{m5} and Favorita~\citep{favorita}. These are our largest datasets with 3060 and 4471 total time-series in their respective hierarchices. For both these datasets, we use the product hierarchy i.e each leaf time-series corresponds to the sales of an item aggregated across the stores. The other datasets include: Tourism-L~\citep{tourism, wickramasuriya2019optimal} which is a dataset consisting of tourist count data; Labour~\citep{labour}, consisting of monthly employment data; Traffic~\citep{cuturi}, consisting of daily occupancy rates of cars on freeways; and Wiki2~\citep{wiki}, consisting of daily views on Wikipedia articles. For Tourism-L we benchmark on both the (Geo)graphic and (Trav)el history based hierarchy. More details about the dataset and the features used for each dataset can be found in Appendix~\ref{sec:dataset_details} and Table~\ref{tab:datastat}. 

\begin{table}[ht!]
\caption{\small We present the normalized metrics WAPE / NRMSE across all levels. We report normalized metrics so that they can be averaged to produce the mean column. All the numbers are averaged over 5 runs. The numbers in bold represent the statistically significant best performances in each column, while the italized ones represent the second best.\\}
\label{tab:point}
\centering
\resizebox{\linewidth}{!}{%
\begin{tabular}{l|p{70pt}|c|c|c|c|c}
\toprule
Dataset & Method & L0 & L1 & L2 & L3 & Mean \\
\midrule
\multirow{6}{*}{M5} & \oursp & \textbf{0.0404 / 0.0487} & \textbf{0.0518 / 0.0773} & \textbf{0.0662 / 0.1061} & \textbf{0.3238 / 0.6818} & \textbf{0.1205 / 0.2285} \\
 & Fedformer-Base (incoherent) & \textit{0.0585 / 0.0715} & \textit{0.0659 / 0.1036} & \textit{0.0718 / 0.1219} & 0.3453 / 0.7065 & \textit{0.1354 / 0.2509} \\
 & Fedformer-BU & 0.0641 / 0.0781 & 0.0739 / 0.1167 & 0.0795 / 0.1167 & 0.3453 / 0.7065 & 0.1407 / 0.2603 \\
 & Fedformer-TD & \textit{0.0585 / 0.0715} & 0.0680 / 0.1067  & 0.0738 / 0.1258  & \textit{0.3443 / 0.699} & \textit{0.1361 / 0.2507}\\
 & Fedformer-ERM & 0.0979 / 0.1145 & 0.102 / 0.1462 & 0.1101 / 0.1755  & 0.3914 / 0.8100 &  0.1753 / 0.3110 \\
 & Fedformer-MinT & 0.0619 / 0.0749  & 0.0714 / 0.1131 & 0.0772 / 0.136  & 0.3462 / 0.7058  & 0.1392 / 0.2575 \\
 \midrule
\multirow{6}{*}{Favorita} & \oursp & \textbf{0.0485 / 0.0614}  & \textbf{0.0948 / 0.2336}  & \textit{0.1513 / 0.4504} & \textbf{0.3039 / 1.0925}  & \textbf{0.1496 / 0.4595} \\
 & Fedformer-Base (incoherent) &  \textit{0.0667 / 0.0869} & \textit{0.1004 / 0.2562} & 0.1904 / 0.4447  & 0.3875 / 1.1264   &  0.1863 / 0.4786 \\
 & Fedformer-BU & 0.0887 / 0.101  & 0.1114 / 0.2848  & 0.1605 / 0.4551 & 0.3875 / 1.1264  &  0.187 / 0.4918 \\
 & Fedformer-TD & \textit{0.0667 / 0.0869} & \textit{0.098 / 0.2492} & 0.2482   / 2.4286 & 0.5343   / 7.3585  & 0.2368 / 2.5308 \\
 & Fedformer-ERM & 0.0746 / 0.0967 & \textit{0.0994 / 0.2721} & \textbf{0.1385 / 0.4397}  & \textit{0.3116 / 1.1243} & \textit{0.156 / 0.4832} \\
 & Fedformer-MinT & \textit{0.0667 / 0.0869} & 0.1035 / 0.2541 & 0.1777 / 0.4403 & 0.4089 / 1.1309 & 0.1892 / 0.478 \\
\bottomrule
\end{tabular}%
}
\end{table}

Note that for the sake of reproducibility, except for the additional M5 and Favorita datasets, the datasets and experimental setup are largely identical to that in \citep{rangapuram2021end} with an increased horizon for traffic and wiki datasets. In~\citep{rangapuram2021end}, the prediction window for the latter two datasets were chosen to be only 1 time-step which is extremely small; moreover on traffic the prediction window only includes the day Dec 31st which is atypical especially because the dataset includes only an year of daily data. Therefore we decided to increase the validation and test size to 7. In all datasets the last $F$ time-steps form the test window while the preceding $F$ time-steps form the validation window -- the same convention was followed in~\citep{rangapuram2021end}.

\begin{table}[ht!]
\caption{\small Normalized CRPS scores for M5 and Favorita. We average the deep learning based methods over 5 independent runs. The rest of the methods had very little variance. We report the corresponding standard error and only bold numbers that are the statistically significantly better than the rest. The second best numbers in each column are italicized.\\}
\label{tab:main_results}
\centering
\resizebox{\linewidth}{!}{%
\begin{tabular}{p{80pt}|c|c|c|c|c}
\toprule
M5      & L0 & L1 & L2 & L3 & Mean \\ \midrule
\oursp  &
\B{0.0379}{0.0014} &
\B{0.0422}{0.0004} &
\B{0.0536}{0.0023} &
\B{0.2543}{0.0067} &
\B{0.0970}{0.0013} \\[2pt]
\hier  &
\R{0.1129}{0.0008} &
\R{0.1106}{0.0008} &
\R{0.1167}{0.0010} &
\R{0.2940}{0.0012} &
\R{0.1586}{0.0005} \\[2pt]
\permbu  &
0.0639 &
0.0673 &
0.0737 &
0.2978 &
0.1257 \\[2pt]
Best of Nixtla (AutoARIMA-TD) &
\textit{0.0599} &
{\it 0.0643} &
{\it 0.0713} &
{\it 0.2808} &
{\it 0.1191} \\
\bottomrule
\end{tabular}%
}

\centering
\resizebox{\linewidth}{!}{%
\begin{tabular}{p{80pt}|c|c|c|c|c}
\toprule
Favorita     & L0 & L1 & L2 & L3 & Mean \\ \midrule
\oursp &
\B{0.0430}{0.0024} &
\B{0.0709}{0.0016} &
\B{0.1132}{0.0017} &
\B{0.2446}{0.0023} &
\B{0.1179}{0.0018} \\[2pt]
\hier &
\R{0.0955}{0.0009} &
\R{0.1211}{0.0018} &
\R{0.1648}{0.0039} &
\R{0.3305}{0.0060} &
\R{0.1780}{0.0028} \\[2pt]
\permbu  &
\textit{0.0561} &
 0.8279 &
0.6142 &
{\it 0.3184} &
0.4541 \\[2pt]
Best of Nixtla (AutoARIMA-BU)  &
\textit{0.0563} &
{\bf 0.0697} &
{\bf 0.1119} &
{\it 0.3190} &
{\it 0.1392} \\[2pt]
\bottomrule
\end{tabular}%
}
\end{table}

\begin{table}[ht!]
\caption{\small Normalized CRPS scores averaged over all levels for all remaining datasets introduced in Sec~\ref{sec:expt}. The full set of level-wise metrics can be found in Appendix~\ref{app:fullresults}. We report the corresponding standard error and only bold numbers that
are the statistically significantly better than the rest. The second best numbers in each column are italicized.\\}
\label{tab:mean_results}
\centering
\begin{tabular}{l|c|c|c|c}
\toprule
Mean metrics & Labour & Traffic & Wiki2 & Tourism-L \\ \midrule
\oursp &
\B{0.0250}{0.0015} &
\B{0.0526}{0.0028} &
\B{0.2706}{0.0048} &
\textbf{0.1407}\\[2pt]
\hier &
\C{0.0340}{0.0088} &
\B{0.0506}{0.0011} &
\C{0.2769}{0.004 } &
0.1520 \\[2pt]
\permbu &
0.0393 &
0.1019 &
0.5033 &
0.2518 \\[2pt]
Best of Nixtla &
{\it 0.0346} (ERM) &
{\it 0.0757} (TD) &
0.3631 (TD) &
\textit{0.1474} (MinT) \\
\bottomrule
\end{tabular}

\end{table}

\textbf{Benefits of end to end hierarchical modeling.} Before comparing our method with state-of-the-art probabilistic coherent forecasting baselines on the benchmark datasets, we would like to showcase the benefit of end-to-end hierarchical modeling as a whole. To that end, we first choose a simpler task of accurate point forecasting on our two largest datasets, Favorita and M5. 

In our baselines, as a base forecaster, we choose a recently published strong multivariate point forecasting method, FEDformer~\citep{zhou2022FEDformer}, that uses a frequency-enhanced transformer (along with other techniques like separate modeling of seasonality and trend) to achieve state-of-the-art results in several long-horizon forecasting tasks. Then we reconcile these base forecasts using popular reconciliation techniques like Bottom-Up~\citep{hyndman2011optimal}, Top-Down~\citep{hyndman2011optimal}, ERM~\cite{ben2019regularized}, MinT~\cite{wickramasuriya2020optimal} to yield the corresponding baselines FEDformer-BU, FEDformer-TD, FEDformer-ERM and FEDformer-MinT. Lastly we also report the metrics for the incoherent base forecasts dubbed FEDformer-Base. We use the open source package Nixtla~\citep{olivares2022hierarchicalforecast} for the reconciliation and the FEDformer repository for the base forecasts.

We present the WAPE / NRMSE metrics (defined in Appendix~\ref{sec:dataset_details}) for the baseline forecasts and our p50 forecasts in Table~\ref{tab:point}. Our end to end coherent method achieves better performance across all levels compared to the base forecasts FEDformer-Base, even though we use a relatively simple architecture. Post-hoc reconciliation seems to help in some cases. For example, FEDformer-ERM has better performance on the Favorita dataset than the base forecasts, but even that falls short of our model (except in L2 of Favorita). This suggests that using coherence as an inductive bias during training might be important in propagating higher level signals to leaf levels. We provide details about all our hyperparameters in Appendix~\ref{app:model}.

{\bf Probabilistic Forecasting.~}
Now that we have seen the benefits of end-to-end hierarchical modeling, we are ready to present our main empirical results for probabilistic hierarchical forecasting. We compare our models to state-of-the-art strictly coherent probabilistic forecasting baselines. The first two baselines capture dependencies using the tree structure during generating the initial probabilistic forecasts even before reconciliation:
(i) \hier~\citep{rangapuram2021end} is an end-to-end deep-learning approach for coherent probabilistic forecasts. This method by design produces coherent probabilistic forecasts.
(ii) \permbu~\citep{taieb2017coherent} is a copula based approach for producing probabilistic hierarchical forecasts. The copula is used to capture dependencies among each family and the the samples are reconciled using well-known reconciliation methods like MinT and BottomUP (BU). We report the best numbers between \permbu-MinT and \permbu-BU. (iii) For the sake of completeness we also include post-hoc reconciliation baselines. We use the Nixtla package~\citep{olivares2022hierarchicalforecast} that produces base probabilistic forecasts from \href{https://www.statsmodels.org/dev/generated/statsmodels.tsa.arima.model.ARIMA.html}{AutoARIMA} and then uses MinT, Bottom-Up (BU), TopDown (TD), and ERM reconciliation using a reverse engineered empirical covariance matrix to provide probabilistic forecasts. (Note that we cannot use Fedformer for base forecasts here, since it does not generate probabilistic forecasts.) In the interest of space, we only provide the numbers for the best performing Nixtla method in Table~\ref{tab:main_results}, while providing the detailed numbers in the~ {Appendix}.

\textbf{Evaluation.} We evaluate forecasting accuracy using the continuous ranked probability score (CRPS). The CRPS is minimized when the predicted quantiles match the true data distribution~\citep{gneiting2007strictly}. This is the standard metric used to benchmark probabilistic forecasting in numerous papers~\citep{rangapuram2018deep, rangapuram2021end,taieb2017coherent}. Similar to \citet{rangapuram2021end}, we also normalize the CRPS scores at each level, by the absolute sum of the true values of all the nodes of that level. 
We mathematically define the CRPS score in Appendix~\ref{app:fullresults}.

 We present level-wise performance of all methods on M5 and Favorita, as well as mean performance  on the other datasets (the full level-wise metrics on all datasets can be seen in Appendix \ref{app:fullresults}). In these tables, we highlight in bold numbers that are statistically
significantly better than the rest. The second best numbers in each column are italicized. The deep learning based methods are averaged over 5 runs while other methods had very little variance. 

{\bf M5:}  
We see that overall in all columns, \oursp\ performs much better than all the baselines (around $18$\% better than the best baseline (AutoARIMA-TD) on the mean). Interestingly, even a simple top-down baseline like AutoARIMA-TD outperforms recent state-of-the-art models like \hier~and \permbu, attesting to the power of top-down approaches.
 We hypothesize that \hier\ does not work well on these larger datasets because the DeepVAR model needs to be applied to thousands of time-series, which leads to a prohibitive size of the fully connected input layer and a hard joint optimization problem.

 {\bf Favorita:} As in the previous results, even in Favorita, \oursp\ outperforms the other models by a large margin, resulting in a $15$\% better mean performance than the best baseline. Interestingly again, a simple AutoARIMA-BU model outperforms both \hier~and \permbu.

 {\bf Other Datasets:}  Table~\ref{tab:mean_results} presents mean CRPS scores on Labor, Traffic, Tourism and Wiki2 datasets. In Traffic, \oursp~ is within statistical error of the best baseline, but in the other datasets, \oursp\ comfortably outperforms all baselines. In three of the four smaller datasets, \hier~ performs better than the other baselines, which suggest that \hier~works reasonably well on smaller datasets compared to reconciliation-based methods (though  \oursp~ is still significantly better on Labor, Wiki2, and Tourism by 26\%, 2\% and 4.5\% respectively. We provide more detailed results for all these datasets in Appendix~\ref{app:fullresults}.
 
\begin{table}[ht!]
\caption{\small We provide an ablation study in the Favorita dataset. We only provide the mean forecast for the sake of brevity but our original model outperforms the ablated baselines on all levels.}
\label{tab:ablation}
\centering
\begin{tabular}{l|c}
\toprule
Favorita Ablation & Mean \\ \midrule
\oursp &
\B{0.1179}{0.0018} \\[2pt]
\oursnoatt &
\R{0.1340}{0.0028}  \\[2pt]
\historical &
\R{0.1436}{0.0010}
\\
\bottomrule
\end{tabular}

\end{table}

{\bf Ablation.} In Table~\ref{tab:ablation}, we study the role of various components in our model on the Favorita dataset. We first remove the attention layers after the encoder and see a $13\%$ drop in mean metric. Thus mixing the information among the children in important. We also use historical static fractions in combination to the root (L0) predictions from our model. It can be seen that our learnt proportions model outperforms historical proportions.

\section{Conclusion}
\label{sec:conclusion}
In this paper, we proposed a probabilistic top-down based hierarchical forecasting approach, that obtains coherent, probabilistic forecasts without the need for a separate reconciliation stage. Our approach is built around a novel deep-learning model for learning the distribution of  proportions according to which a parent time series is disaggregated into its children time series. We show in empirical evaluation on several public datasets, that our model obtains state-of-the-art results compared to previous methods.

For future work, we plan to explore extending our approach to handle more complex hierarchical structural constraints, beyond trees. We would also like to note that currently our theoretical justification only applies to learning historical proportions; it would be interesting to extend it to predicted future proportions.

\FloatBarrier
\clearpage
\bibliography{new_main}

\clearpage
\appendix
\onecolumn

\section{Related Work on Hierarchical Forecasting}
\label{app:rwork}
{\bf Coherent Point Forecasting:~} As mentioned earlier, many existing coherent hierarchical forecasting methods rely on a two-stage reconciliation approach. More specifically, given non-coherent base forecasts $\hat{\by}_t\in \reals^N$, reconciliation approaches aim to design a projection matrix $\bP \in \reals^{m \times N}$ that can project the base forecasts linearly into new leaf forecasts, which are then aggregated using $\bS$ to obtain (coherent) revised forecasts $\tilde{\by}_t = \bS\bP\hat{\by}_t\in \reals^N$. The post-processing is call reconciliation or on other words base forecasts are reconciled.
Different hierarchical methods specify different ways to optimize for the $\bP$ matrix. The naive Bottom-Up  approach~\citep{hyndman2018forecasting} simply aggregates up from the base leaf predictions to obtain revised coherent forecasts. The MinT method~\citep{wickramasuriya2019optimal}
computes $\bP$ that obtains the minimum variance unbiased revised forecasts, assuming unbiased base forecasts. The ERM method from ~\cite{ben2019regularized} optimizes $\bP$ by directly performing empirical risk minimization over the mean squared forecasting errors. Several other criteria~\citep{hyndman2011optimal, van2015game, panagiotelis2020probabilistic} for optimizing for $\bP$ have also been proposed. Note that some of these reconciliation approaches like MinT can be used to generate confidence intervals by estimating the empirical covariance matrix of the base forecasts~\cite{hyndman2018forecasting}.

{\bf Coherent Probabilistic Forecasting:} The PERMBU method in ~\cite{taieb2017coherent} is a reconciliation based hierarchical approach for probabilistic forecasts. It starts with independent marginal probabilistic forecasts for all nodes, then uses samples from marginals at the leaf nodes, applies an empirical copula, and performs a mean reconciliation step to obtain revised (coherent) samples for the higher level nodes. \citet{athanasopoulos2020hierarchical} also discuss two approaches for coherent probabilistic forecasting: (i) using the empirical covariance matrix under the Gaussian assumption and (ii) using a non-parametric bootstrap method.
The recent work of \citet{rangapuram2021end} is a single-stage end-to-end method that uses deep neural networks to obtain coherent probabilistic hierarchical forecasts. Their approach is to use a neural-network based multivariate probabilistic forecasting model to jointly model all the time series and explicitly incorporate a differentiable reconciliation step as part of model training, by using sampling and projection operations. A recent approach of \citet{olivares2021probabilistic} uses a Deep Poisson Mixture Network to 
models the joint probability of the leaf time series as a finite mixture of Poisson distributions. 

{\bf Approximately Coherent Methods:~} Several approximately-coherent hierarchical models have also been recently proposed, that mainly use the hierarchy information for improving prediction quality, but do not guarantee strict coherence, and often do not generate probabilistic predictions. Many of them \citep{mishchenko2019self, gleason2020forecasting, han2021simultaneously, han2021mecats, paria2021hierarchically} use regularization-based approaches to incorporate the hierarchy tree into the model via $\ell_2$ regularization.  \citet{kamarthi2022profhit} imposes approximate coherence on probabilistic forecasts via regularization of the output distribution.

\section{Proofs}
\subsection{Proof of Theorem~\ref{thm:risk-ratio}}\label{sec:proof-risk-ratio}
We prove the claims about the excess risk of top-down and bottom-up approaches in the following two sections. Recall that ordinary least square (OLS) estimator $\hat{\theta} = \bb{\sum_{i=1}^n\x_i\x_i^\top}^{-1}\sum_{i=1}^n\x_iy_i$. The population squared error of a linear predictor is defined as $(\hat{\theta}-\theta)^\top\Sigma(\hat{\theta}-\theta)$, which is also known as excess risk. 
\subsubsection{Excess risk of the top down approach}
For the root node, the OLS predictor is written as 
$$
\hat{\theta}_0 = \bb{\sum_{t=1}^n\x_t\x_t^\top}^{-1}\sum_{t=1}^n\x_ty_{t,0},
$$
and the expected excess risk is
\begin{align}
&\E[(\hat\theta_0-\theta_0)^\top\Sigma(\hat\theta_0-\theta_0)]\nonumber\\
\stackrel{\text{(a)}}{=}\;& \E\sbb{\sigma^2\sum_{ {t}=1}^n\x_t^\top\bb{\sum_{ {r=1}}^{ {n}} \x_t\x_t^\top}^{-1}\Sigma\bb{\sum_{ {r=1}}^{ {n}} \x_t\x_t^\top}^{-1}\x_t}\nonumber\\
\stackrel{\text{(b)}}{=}\;& \Tr\sbb{\E\sbb{\sigma^2\bb{\sum_{ {t=1}}^{ {n}} \x_t\x_t^\top}^{-1}\Sigma}}\nonumber\\
\stackrel{\text{(c)}}{=}\;& \sigma^2 d/(n-d-1)\label{eqn:risk-root}
\end{align}
where equation (a) holds by expanding $y_{i,0} = \x_i^\top\theta_0+\eta_i$ and the fact that $\eta_i$ is independent of $\x_i$, equation (b) holds by the property of trace, and equation (c) follows from the mean of the inverse-Wishart distribution.

For each children node, we learn the proportion coefficient with 
$$
\hat p_i = \frac{1}{n}\sum_{t=1}^n\frac{y_{t,i}}{y_{t,0}}.
$$
Notice that
\begin{align}
\Var[\hat p_i] &= \frac{1}{n}\Var\sbb{\frac{y_{1,i}}{y_{1,0}}}\nonumber\\
&= \frac{1}{n}\Var[a_i] \nonumber\\
&= s_i/n\label{eqn:varqi}.
\end{align}
Recall that the optimal linear predictor of the $i$-th child node is $p_i\theta_0$. Therefore, the expected excess risk of the top down predictor is 
\begin{align*}
 &\E\sbb{(\hat{p}_i \hat\theta_0 - p_i\theta_0 )^\top\Sigma (\hat{p}_i \hat\theta_0 -p_i \theta_0)}\\
=\; &\E\sbb{(\hat{p}_i \hat\theta_0 - {p}_i \hat\theta_0 + {p}_i \hat\theta_0-p_i\theta_0)^\top\Sigma (\hat{p}_i \hat\theta_0 - {p}_i \hat\theta_0 + {p}_i \hat\theta_0-p_i\theta_0)}\\
=\; &\E\sbb{(\hat p_i - p_i)^2\hat\theta_0^\top\Sigma\hat\theta_0 + p_i^2 (\hat\theta_0-\theta_0)^\top\Sigma(\hat\theta_0-\theta_0)}\\
\stackrel{\text{(a)}}{=}\;&\frac{1}{n}s_i\theta_0^\top\Sigma\theta_0 + \bb{\frac{1}{n}s_i+p_i^2}\frac{d}{n-d-1}\sigma^2\label{eqn:risk-topdown},
\end{align*}
where we have applied Equation~\ref{eqn:varqi} and Equation~\ref{eqn:risk-root} in equality (a). Taking summation over all the children, we get the total excess risk equals
$$
\frac{\sum_{i=1}^K{s_i}}{n}\theta_0^\top\Sigma\theta_0+\bb{\frac{\sum_{i=1}^K s_i}{n}+\sum_{i=1}^Kp_i^2}\frac{d}{n-d-1}\sigma^2
$$

\subsubsection{Excess risk of the bottom up approach}
For the $i$-th child node, the OLS estimator is 
$$
\hat\theta_i = \bb{\sum_{t=1}^n \x_t\x_t^\top}^{-1}\sum_{t=1}^n \x_t y_{t,i}.
$$
Recall that the best linear predictor of the $i$-th child node is $p_i\theta_0$
The excess risk is 
\begin{align*}
&\E(\hat\theta_i-p_i\theta_0)^\top\Sigma(\hat\theta_i-p_i\theta_0)\\
=\; & \E\sbb{((\hat\theta_i-p_i\hat\theta_0 ) + (p_i\hat\theta_0 -p_i\theta_0))^\top\Sigma((\hat\theta_i-p_i\hat\theta_0) + (p_i\hat\theta_0 - p_i\theta_0))}
\end{align*}
Notice that the cross term has $0$ expectation as
\begin{align*}
&\E[(\hat\theta_i-p_i\hat\theta_0)^\top\Sigma(p_i\hat\theta_0-p_i\theta_0)]\\
\stackrel{\text{(a)}}{=}\;& \E\sbb{\E_{\a}\sbb{\sum_{t=1}^n (a_{t,i} y_{t,0} -p_iy_{t,0})\x_t^\top \bb{\sum_{t=1}^n \x_t\x_t^\top}^{-1}\Sigma(p_i\hat\theta_0-p_i\theta_0)}}\\
\stackrel{\text{(b)}}{=}\;& 0,
\end{align*}
where the first equality holds by the definition of node $i$-th value $y_{t,i}$. Therefore, it holds that
\begin{align*}
& \E\sbb{((\hat\theta_i-p_i\hat\theta_0 ) + (p_i\hat\theta_0 -p_i\theta_0))^\top\Sigma((\hat\theta_i-p_i\hat\theta_0) + (p_i\hat\theta_0 - p_i\theta_0))}\\
=\;& \E\sbb{(\hat\theta_i-p_i\hat\theta_0)^\top\Sigma(\hat\theta_i-p_i\hat\theta_0)} + p_i^2\frac{d}{n-d-1}\sigma^2\\
=\;& \E\Tr\sbb{\sum_{j=1}^n (a_{j,i}-p_{i})^2y_{t,0}^2\x_j\x_j^\top \bb{\sum_{t=1}^n \x_t\x_t^\top}^{-1}\Sigma\bb{\sum_{t=1}^n \x_t\x_t^\top}^{-1}}+ p_i^2\frac{d}{n-d-1}\sigma^2\\
=\;& s_i \E\Tr\sbb{\sum_{j=1}^n \left((\theta_0^\top\x_j)^2+\eta_j^2\right)\x_j\x_j^\top \bb{\sum_{t=1}^n \x_t\x_t^\top}^{-1}\Sigma\bb{\sum_{t=1}^n \x_t\x_t^\top}^{-1}}+ p_i^2\frac{d}{n-d-1}\sigma^2\\
\stackrel{\text{(a)}}{\ge}\;&s_i\sigma^2\E\Tr\sbb{\bb{\sum_{j=1}^n \x_j\x_j^\top} \bb{\sum \x_i\x_i^\top}^{-1}\Sigma\bb{\sum \x_i\x_i^\top}^{-1}}+p_i^2\frac{d}{n-d-1}\sigma^2\\
\stackrel{(b)}{=}\;& (s_i +p_i^2)\sigma^2 \frac{d}{n-d-1},
\end{align*}
where inequality (a) holds since $(\theta^\top_0\x_j)^2$ term is non-negative, equality (b) holds by the property of inverse-Wishart distribution. 
Taking summation over all the children, we get the total excess risk is lower bounded by 
$$
\sum_{i=1}^K (s_i +p_i^2)\frac{d}{n-d-1} \sigma^2 
$$
This concludes the proof.
\subsection{Proof of Corollary~\ref{cor:diri-risk-ratio}}\label{sec:proof-diri-risk-ratio}
In this section, we apply Theorem~\ref{thm:risk-ratio} to Dirichlet distribution to show that the excess risk of bottom-up approach is $\min(d,K)$ times higher than top-down approach for a natural setting.

Recall that a random vector $\a$ drawn from a $K$-dimensional Dirichlet distribution $\Dir(\alpha)$ with parameters $\alpha$ has mean $\E[\a] = \frac{1}{\sum_{i=1}^K\alpha_i} \alpha$, and the variance $\Var[a_i]=\frac{\alpha_i(1-\alpha_i)}{\sum_{i=1}^K\alpha_i+1}$. Let $\alpha_i=\frac{1}{K}$ for all $i\in [K]$, $\theta_0^\top\Sigma\theta_0=\sigma^2$. The total excess risk of the top-down approach is 
\begin{align*}
\E\sbb{\sum_{i=1}^K r(\hat\theta^{\t}_i)}\;=&\frac{\sum_{i=1}^K{s_i}}{n}\theta_0^\top\Sigma\theta_0+\bb{\frac{\sum_{i=1}^K s_i}{n}+\sum_{i=1}^Kp_i^2}\frac{d}{n-d-1}\sigma^2\\
=\;& \bb{\frac{1-1/K}{2n}+\bb{\frac{1-1/K}{2n}+\frac{1}{K}}\frac{d}{n-d-1}}\sigma^2.
\end{align*}
The total excess risk of the bottom-up approach is lower bounded by
\begin{align*}
    \E\sbb{\sum_{i=1}^K r({\hat\theta}^{\mathbf b}_i )}=\;&(s_i +p_i^2) \frac{d}{n-d-1} \sigma^2\\
    =\;& \bb{\frac{1-1/K}{2}+\frac{1}{K}} \frac{d}{n-d-1}\sigma^2
\end{align*}
Now assuming that $n\ge 2d$, the top-down approach has expected risk $\E[\sum_{i=1}^K r(\hat\theta^{\t}_i)] = O(\frac{1}{n}+\frac{d}{nK})$, and the bottom-up approach has expected risk $\E[\sum_{i=1}^K r(\hat\theta^{\t}_i)] = \Omega(\frac{d}{n})$. Therefore, it holds that
$$
\frac{\E[\sum_{i=1}^K r(\hat\theta^{\mathbf b}_i)]}{\E[\sum_{i=1}^K r(\hat\theta^{\t}_i)]} = \Omega(\min(d, K))
$$

\section{Datasets}
\label{sec:dataset_details}


\begin{table}
\caption{Normalized CRPS scores on all datasets. We average the deep learning based methods over 5 independent runs. The rest of the methods did not show any variance. We report the corresponding standard error and only bold the numbers that are significantly better than the rest. We also report the mean performance across all levels in the corresponding column.
For ease of comparison, we restate some of the PERMBU numbers by \citet{rangapuram2018deep}.\\
~}
\label{tab:full_results1}
\centering
\tiny
\begin{tabular}{p{80pt}|c|c|c|c|c}
\toprule
M5      & L0 & L1 & L2 & L3 & Mean \\ \midrule
\oursp  &
\B{0.0379}{0.0014} &
\B{0.0422}{0.0004} &
\B{0.0536}{0.0023} &
\B{0.2543}{0.0067} &
\B{0.0970}{0.0013} \\[2pt]
\hier  &
\R{0.1129}{0.0008} &
\R{0.1106}{0.0008} &
\R{0.1167}{0.0010} &
\R{0.2940}{0.0012} &
\R{0.1586}{0.0005} \\[2pt]
\permbu  &
0.0639 &
0.0673 &
0.0737 &
0.2978 &
0.1257 \\[2pt]
AutoARIMA-BU &
0.1188 &
0.1173 &
0.1202 &
0.2945 &
0.1627 \\[2pt]
AutoARIMA-ERM &
2.9453 &
3.0110 &
3.0552 &
14.613 &
5.9062 \\[2pt]
AutoARIMA-TD &
0.0599 &
0.0643 &
0.0713 &
0.2808 &
0.1191 \\[2pt]
AutoARIMA-MinT &
0.0566 &
0.0725 &
0.0880 &
0.3074 &
0.1311 \\[2pt]
\bottomrule
\end{tabular}\\[5pt]
\centering
\tiny
\begin{tabular}{p{80pt}|c|c|c|c|c}
\toprule
Favorita     & L0 & L1 & L2 & L3 & Mean \\ \midrule
\oursp &
\B{0.0430}{0.0024} &
\B{0.0709}{0.0016} &
\B{0.1132}{0.0017} &
\B{0.2446}{0.0023} &
\B{0.1179}{0.0018} \\[2pt]
\hier &
\R{0.0955}{0.0009} &
\R{0.1211}{0.0018} &
\R{0.1648}{0.0039} &
\R{0.3305}{0.0060} &
\R{0.1780}{0.0028} \\[2pt]
\permbu  &
0.0561 &
 0.8279 &
0.6142 &
0.3184 &
0.4541 \\[2pt]
AutoARIMA-BU &
0.0563 &
\textbf{0.0697} &
\textbf{0.1119} &
0.3190 &
0.1392 \\[2pt]
AutoARIMA-ERM &
1.4857 &
1.7470 &
2.4220 &
4.8256 &
2.6201 \\[2pt]
AutoARIMA-TD &
0.0802 &
0.2606 &
0.5253 &
1.1120 &
0.4945 \\[2pt]
AutoARIMA-MinT &
0.0781 &
0.1539 &
0.2456 &
0.4448 &
0.2306 \\[2pt]
\bottomrule
\end{tabular}
\\[5pt]
\centering
\resizebox{\textwidth}{!}{
\begin{tabular}{l|c|c|c|c|c|c|c|c|c}
\toprule
Tourism      & L0 & L1 (Geo) & L2 (Geo) & L3 (Geo) &  L1 (Trav) & L2 (Trav) & L3 (Trav) & L4 (Trav) & Mean \\ \midrule
\oursp         &
\R{0.0457}{0.0033} &
\R{0.0823}{0.0027} &
\R{0.1341}{0.0028} &
\B{0.1769}{0.0027} &
\R{0.0812}{0.0018} &
\R{0.1358}{0.0008} &
\B{0.2015}{0.0009} &
\B{0.2684}{0.0008} &
\B{0.1407}{0.0022}
\\[2pt]
\hier &
\R{0.0810}{0.0053} &
\R{0.1030}{0.0030} &
\R{0.1361}{0.0024} &
\B{0.1752}{0.0026} &
\R{0.1027}{0.0062}  &
\R{0.1403}{0.0047} &
\R{0.2050}{0.0028} &
\R{0.2727}{0.0017} &
\R{0.1520}{0.0033}\\[2pt]
PERMBU &
0.131 &
0.129 &
0.1723 &
0.2189 &
0.1698 &
0.3063 &
0.5461 &
0.3415 &
0.2518
\\[2pt]
AutoARIMA-BU &
0.1240 &
0.1166 &
0.1612 &
0.2134 &
0.1249 &
0.1554 &
0.2401 &
0.3428 &
0.1848
\\[2pt]
AutoARIMA-ERM &
0.0465 &
0.1181 &
0.1970 &
0.2781 &
\textbf{0.0678} &
0.1571 &
0.2952 &
0.4304 &
0.1987
\\[2pt]
AutoARIMA-TD &
0.0332 &
0.0823 &
0.1547 &
0.2137 &
0.4237 &
0.7107 &
1.0285 &
1.2487 &
0.4869
\\[2pt]
AutoARIMA-MinT &
\textbf{0.0322} &
\textbf{0.0673} &
\textbf{0.1270} &
0.1999 &
0.0733 &
\textbf{0.1298} &
0.2149 &
0.3352 &
0.1474
\\
\bottomrule
\end{tabular}
}
\\[5pt]
\centering
\tiny
\begin{tabular}{l|c|c|c|c|c}
\toprule
Labour & L0 & L1 & L2 & L3 & Mean \\ \midrule
\oursp &
\B{0.0172}{0.0019}&
\B{0.0242}{0.0018}&
\B{0.0243}{0.0016}&
\B{0.0345}{0.0007}&
\B{0.0250}{0.0015} \\[2pt]
\hier &
\R{0.0311}{0.0120} &
\R{0.0336}{0.0089} &
\R{0.0336}{0.0082} &
\R{0.0378}{0.0060} &
\R{0.0340}{0.0088}\\[2pt]
PERMBU &
0.0406 &
0.0389 &
0.0382 &
0.0397 &
0.0393
\\[2pt]
AutoARIMA-BU &
0.0314 &
0.0402 &
0.0393 &
0.0361 &
0.0368 \\[2pt]
AutoARIMA-ERM &
0.0246 &
0.0306 &
0.0335 &
0.0495 &
0.0346 \\[2pt]
AutoARIMA-TD &
0.0343 &
0.0458 &
0.0462 &
0.0462 &
0.0431 \\[2pt]
AutoARIMA-MinT &
0.0396 &
0.0392 &
0.0410 &
0.0436 &
0.0409 \\[2pt]

\bottomrule
\end{tabular}
\\[5pt]
\centering
\tiny
\begin{tabular}{l|c|c|c|c|c}
\toprule
Traffic ($F=7$) & L0 & L1 & L2 & L3 & Mean \\ \midrule
\oursp &
\B{0.0213}{0.0041} &
\B{0.0247}{0.0039} &
\B{0.0296}{0.0032} &
\R{0.1350}{0.0001} & 
\R{0.0527}{0.0028} \\[2pt]
\hier &
\R{0.0245}{0.0011} &
\B{0.0268}{0.001 } &
\B{0.0307}{0.0011} &
\B{0.1206}{0.0019} &
\B{0.0506}{0.0011} \\[2pt]
PERMBU &
0.0780 &
0.0744 &
0.0708 &
0.1844 &
0.1019 \\[2pt]
AutoARIMA-BU &
0.0682 &
0.0648 &
0.0621 &
0.1832 &
0.0946 \\[2pt]
AutoARIMA-ERM &
0.0997 &
0.1086 &
0.1117 &
0.3364 &
0.1641 \\[2pt]
AutoARIMA-TD &
0.0486 &
0.0507 &
0.0549 &
0.1485 &
0.0757 \\[2pt]
AutoARIMA-MinT &
0.0340 &
0.0429 &
0.0570 &
0.1859 &
0.0800
\\
\bottomrule
\end{tabular}
\\[5pt]
\centering
\tiny
\begin{tabular}{l|c|c|c|c|c|c}
\toprule
Wiki2 ($F=7$) & L0 & L1 & L2 & L3 & L4 & Mean \\ \midrule
\oursp &
\B{0.1483}{0.0116} &
\B{0.2096}{0.0056} &
\B{0.2817}{0.0042} &
\B{0.29}{0.0036} &
\B{0.4233}{0.005} &
\B{0.2706}{0.0048} \\[2pt]
\hier &
\B{0.133 }{0.0102} &
\B{0.2094}{0.0057} &
\R{0.2942}{0.0032} &
\R{0.3057}{0.0031} &
\R{0.4421}{0.0016} &
\R{0.2769}{0.004 } \\[2pt]
PERMBU &
0.1859 &
0.3437 &
0.5551 &
0.5635 &
0.8685 &
0.5033 \\[2pt]
AutoARIMA-BU &
0.1954 &
0.3853 &
0.6083 &
0.6155 &
0.9732 &
0.5555 \\[2pt]
AutoARIMA-ERM &
0.3238 &
0.4981 &
0.6285 &
0.6458 &
0.9778 &
0.6148 \\[2pt]
AutoARIMA-TD &
0.2449 &
0.3398 &
0.3841 &
0.389 &
0.4577 &
0.3631 \\[2pt]
AutoARIMA-MinT &
0.2171 &
0.3651 &
0.5525 &
0.6542 &
1.2531 &
0.6084
\\
\bottomrule
\end{tabular}

\end{table}


We use publicly available benchmark datasets for our experiments.
\begin{enumerate}[leftmargin=15pt,nolistsep,itemsep=4pt]
    \item M5 \footnote{\url{https://www.kaggle.com/c/m5-forecasting-accuracy/}}: It consists of time series data of product sales from 10 Walmart stores in three US states. The data consists of two different hierarchies: the product hierarchy and store location hierarchy. For simplicity, in our experiments we use only the product hierarchy consisting of 3k nodes and 1.8k time steps. Time steps 1907 to 1913 constitute a test window of length 7. Time steps 1 to 1906 are used for training and validation.
    \item Favorita \footnote{\url{https://www.kaggle.com/c/favorita-grocery-sales-forecasting/}}: It is a similar dataset, consisting of time series data from Corporaci\'on Favorita, a South-American grocery store chain. As above, we use the product hierarchy, consisting of 4.5k nodes and 1.7k time steps. Time steps 1681 to 1687 constitute a test window of length 7. Time steps 1 to 1686 are used for training and validation.
    \item Tourism-L\footnote{\url{https://robjhyndman.com/publications/mint/}}: consists of monthly domestic tourist count data in Australia across 7 states which are sub-divided into regions, sub-regions, and visit-type. The data consists of around 500 nodes and 228 time steps. This dataset consists of two hierarchies (Geo and Trav) as also followed in \citep{rangapuram2021end}. Time steps 1 to 221 are used for training and validation. The test metrics are computed on steps 222 to 228.
    
    \item Traffic \citep{cuturi}: Consists of car occupancy data from freeways in the Bay Area, California, USA. The data is aggregated in the same way as \citep{ben2019regularized}, to create a hierarchy consisting of 207 nodes spanning 366 days. Time steps 1 to 359 are used for training and validation. The remaining 7 time steps are used for testing.
    
    \item Labour: Australian employement data consisting of 514 time steps sampled monthly, and 57 node hierarchy.
    
    \item Wiki2: This dataset is derived from a larger dataset consisting of daily views of 145k Wikipedia articles. We use a smaller version of the dataset introduced by \citet{ben2019regularized} which consists of a subset of 150 bottom level time series, and 199 total time series.
\end{enumerate}
For both M5 and Favorita we used time features corresponding to each day including day of the week and month of the year. We also used holiday features, in particular the distance to holidays passed through a squared exponential kernel. In addition, for M5 we used features related to SNAP discounts, and features related to oil prices for Favorita. For Tourism, Traffic, Labour, and Wiki2 we only used date features such as day of the week, month of the year, and holiday features from the GluonTS package \citep{gluonts_jmlr}. All the input features were normalized to -0.5 to 0.5.

\begin{table}[ht!]
\caption{Dataset characteristics. $F$ denotes the horizon.}
\label{tab:datastat}
\centering
\begin{tabular}{l|c|c|c|c|c}
\toprule
Dataset      & Total time series & Leaf time series & Levels & Observations & $F$ \\ \midrule
M5 &
3060 &
3049 &
4 &
1913 &
35 days \\[2pt]
Favorita &
4471 &
4100 &
4 &
1687 &
35 days \\[2pt]
Tourism-L (Geo) &
111 &
76 &
4 &
228 &
12 months \\[2pt]
Tourism-L (Trav) &
445 &
304 &
5 &
228 &
12 months \\
Traffic &
207 &
200 &
4 &
366 & 
7 days \\
Labour &
57 &
32 &
4 &
514 & 
8 months \\
Wiki2 &
199 &
150 &
5 &
366 & 
7 days \\
\bottomrule
\end{tabular}%

\end{table}

{\bf Metrics:} For point forecasting we use the metrics WAPE (Weighted Average Percentage Error) and NRMSE (Normalized Root Mean Squared Error). The definitions are as follows:

\begin{align*}
    \mathrm{wape}\left(\bY, \hat{\bY}\right) &= \frac{\sum_{t, i}|Y_{t, i}  - \hat{Y}_{t, i}|}{ \sum_{t, i}|Y_{t, i}|} \\
    \mathrm{nrmse}\left(\bY, \hat{\bY}\right) &= \frac{\sqrt{1/(T*N)\sum_{t, i}(Y_{t, i}  - \hat{Y}_{t, i})^2}}{ 1/(T*N) \sum_{t, i}|Y_{t, i}|}.
\end{align*}

For probabilistic forecasting we use the CRPS metric. Denote the $F$ step $q$-quantile prediction for time series $i$ by $\hat{\bQ}\idx{i}_{\cF}(q) \in \reals^{F}$. $\hat{\bQ}\idx{i}_{s}(q)$ denotes the $q$-th quantile prediction for the $s$-th future time-step for time-series $i$, where $s \in [F]$. Then the CRPS loss is:

\begin{align*}
    &\crps(\hat{\bQ}\idx{i}_\cF(q), \bY\idx{i}_\cF) = \\
    &\frac{1}{F} \sum_{s \in [F]}\int_0^1 2(\II[\bY\idx{i}_s \le \hat{\bQ}\idx{i}_s(q)] - q)(\hat{\bQ}\idx{i}_s(q) - \hat{\bY}\idx{i}_s) dq.
\end{align*}

We normalize the score by the absoluted true values.

\section{Full Results on All Datasets}
\label{app:fullresults}
Tables~\ref{tab:full_results1} show the full set of results for all remaining datasets for all probabilistic baselines.

\section{Additional Experimental Details}
\label{app:model}

{\bf Hyper-parameters and validation.} As mentioned before, we use the last $F$ time-points as the test set and the  $F$ time-points before the test window as the validation set. All hyper-parameters are tuned using the validation set. Then a model with the best hyperparameter (hparams) is trained on training + validation set. We report the metrics obtained by this model on the test set.

In order to reduce the total number of hparams we use a single hparam \nicett{hiddenSize} for all hidden state dimension parameters. This controls the size of hidden state in $\enc$, $\deci$,$\decf$, $\decip$, $\decfp$ and the fully connected layer after each attention layer in $\att$. We tuned this between [128, 256, 512]. The number of hidden layers in $\enc$ is dubbed \nicett{numEncoderLayers} and the number of hidden layers in $\deci$ and $\decip$ is controlled by \nicett{numDecoderLayers}. Both of them were tuned within [2, 3]. The other decoders have only one hidden layer. Learning rate is controlled by \nicett{learningRate}, which was tuned in log-scale from 1e-5 to 1e-2. The attention layer has parameters \nicett{numAttHeads} (tuned in [8, 16, 32]) and \nicett{numAttLayers} (tuned in [2, 3, 5]). We also tune the \nicett{batchSize} within [8, 16, 32].

Now we will specify the chosen hparams for all datasets. 
Note that we also tune the context length among a few values for each dataset similar to what was done in~\citep{rangapuram2021end}.


{\it Favorita.} \nicett{learningRate}: 0.00085, \nicett{hiddenSize}: 128, \nicett{numAttLayers}: 3, \nicett{numAttHeads}: 8, \nicett{batchSize}: 32, \nicett{numEncoderLayers}: 3, \nicett{numDecoderLayers}: 3. The context length is tuned between [140, 70, 35, 28] and 70 was chosen. 

{\it M5.} \nicett{learningRate}: 0.00034, \nicett{hiddenSize}: 128, \nicett{numAttLayers}: 3, \nicett{numAttHeads}: 16, \nicett{batchSize}: 32, \nicett{numEncoderLayers}: 2, \nicett{numDecoderLayers}: 2. The context length is tuned between [140, 70, 35, 28] and 28 was chosen. 

{\it Toursim-L.} \nicett{learningRate}: 0.00007, \nicett{hiddenSize}: 512, \nicett{numAttLayers}: 5, \nicett{numAttHeads}: 16, \nicett{batchSize}: 16, \nicett{numEncoderLayers}: 3, \nicett{numDecoderLayers}: 2. The context length was fixed to 36.

{\it Traffic.} \nicett{learningRate}: 0.0001, \nicett{hiddenSize}: 512, \nicett{numAttLayers}: 2, \nicett{numAttHeads}: 8, \nicett{batchSize}: 16, \nicett{numEncoderLayers}: 3, \nicett{numDecoderLayers}: 3. The context length was fixed to 300.

{\it Labour.} \nicett{learningRate}: 0.00006, \nicett{hiddenSize}: 256, \nicett{numAttLayers}: 3, \nicett{numAttHeads}: 8, \nicett{batchSize}: 16, \nicett{numEncoderLayers}: 3, \nicett{numDecoderLayers}: 2. The context length was tuned in [8, 16, 32 64] and 32 was chosen.

{\it Wiki2.} \nicett{learningRate}: 0.00006, \nicett{hiddenSize}: 512, \nicett{numAttLayers}: 2, \nicett{numAttHeads}: 16, \nicett{batchSize}: 8, \nicett{numEncoderLayers}: 3, \nicett{numDecoderLayers}: 3. The context length was tuned in [140, 70, 35, 28] and 28 was chosen.

{\bf Training details.} Our model is implemented in Tensorflow~\citep{abadi2016tensorflow} and trained using the Adam optimizer with default parameters. We set a step-wise learning rate schedule that decays by a factor of 0.5 a total of 8 times over the schedule. The max. training epoch is set to be 50 while we early stop with a patience of 10. All our experiments were performed on a single server with a 32 core Intel Xeon CPU and an Tesla V100 GPU.

{\bf Baselines.} We used the experimental framework released by \citet{rangapuram2021end} for running the baselines PERMBU, Hier-E2E. On the Favoroita dataset, the original R code of PERMBU does not work because of non positive definite covariance matrix. Therefore we use the implementation in~\cite{olivares2022hierarchicalforecast} as that code is more modular and easy to debug. 

\end{document}